\documentclass[pdflatex,sn-nature]{sn-jnl}

\usepackage{graphicx}%
\usepackage{multirow}%
\usepackage{amsmath,amssymb,amsfonts}%
\usepackage{amsthm}%
\usepackage{mathrsfs}%
\usepackage[title]{appendix}%
\usepackage{xcolor}%
\usepackage{textcomp}%
\usepackage{newunicodechar}
\newunicodechar{−}{\textminus}
\usepackage{manyfoot}%
\usepackage{booktabs}%
\usepackage{algorithm}%
\usepackage{algorithmicx}%
\usepackage{algpseudocode}%
\usepackage{listings}%
\usepackage[most]{tcolorbox} 
\usepackage[most]{tcolorbox}
\usepackage{multicol}   
\newcounter{mybox}
\usepackage{tabularx}
\usepackage{booktabs}
\usepackage[most]{tcolorbox}
\usepackage[table]{xcolor}
\usepackage{tabularx}
\usepackage{colortbl}
\usepackage{xcolor}
\usepackage{caption}
\usepackage{graphicx}
\usepackage{subcaption}
\usepackage{capt-of}
\usepackage{makecell}
\usepackage[section]{placeins} 
\usepackage{hyperref}
\definecolor{headerbg}{RGB}{250, 225, 205}
\usepackage[switch]{lineno} 
\usepackage{float}
\usepackage{placeins}
\usepackage{adjustbox}
\newcolumntype{L}{>{\raggedright\arraybackslash}X}
\newtcolorbox{mhbox}[1][]{
  breakable,
  enhanced,
  colback=gray!12,      
  colframe=gray!12,     
  boxrule=0pt,
  left=1em, right=1em, top=0.6em, bottom=0.9em,
  before skip=10pt, after skip=10pt,
  overlay={
    \draw[line width=0.6pt, black]
      ([xshift=1em,yshift=-2.2em]frame.north west) -- ([xshift=-1em,yshift=-2.2em]frame.north east);
  },
  #1
}

\theoremstyle{thmstyleone}%
\theoremstyle{thmstyletwo}%

\theoremstyle{thmstylethree}%
\newcounter{snote}

\newcounter{edtab}

\usepackage{caption}
\DeclareCaptionLabelSeparator{pipe}{\ \textbar\ }
\makeatletter
\newcommand{\fakelabel}[2]{%
  \protected@write\@auxout{}{\string\newlabel{#1}{{#2}{\thepage}}}%
}
\makeatother

\begin{document}
\fakelabel{supply:icons}{1}

\title[Article Title]{Large language models show systematic failures in incorporating evolving medical knowledge}

\author{\fnm{Min} \sur{Zeng}}

\author{\fnm{Zaifu} \sur{Zhan}}

\author{\fnm{Kai} \sur{Yu}}

\author*{\fnm{Rui} \sur{Zhang}}\email{ruizhang@umn.edu}

\affil[]{\orgdiv{Division of Computational Health Sciences, Department of Surgery}, \orgname{University of Minnesota}, \orgaddress{\city{Minneapolis}, \postcode{55455}, \state{MN}, \country{USA}}}

\unnumbered

\abstract{

Medical knowledge evolves, yet whether large language models (LLMs) faithfully incorporate revisions into clinical recommendations remains unclear. We introduce ChronoEvo, comprising 1,709 recommendation-level revisions across 12 forms of medical knowledge change and 275 guideline lineages, to distinguish recognition of revised guidance from its full incorporation. 
With updated guideline evidence explicitly provided, revision recognition exceeded full incorporation across 14 models, with a 20.1-percentage-point gap in the strongest model. For Qwen3-4B, 90.8\% of responses recognized revisions, 80.2\% were clinically correct, and 58.5\% fully incorporated them, yielding a 32.3-percentage-point gap. Two clinicians' adjudicated assessment of 300 responses yielded a 40.3-percentage-point gap in this enriched validation sample; 61 of 144 responses with incomplete incorporation were clinically incorrect. Prompting Qwen3-4B to prioritize revised evidence reduced outdated-content retention and improved clinical correctness without significantly improving full incorporation. These findings identify faithful incorporation of revised guidance as a distinct challenge for clinical LLM reliability.

}

\maketitle

\section{Introduction}\label{sec1}
Medical knowledge changes as new evidence revises treatment recommendations, patient eligibility, and standards of care. For large language models (LLMs) increasingly considered for clinical decision support~\cite{singhal2023clinical,hager2024clinical,lee2023benefits,tu2025conversational}, reliability therefore depends not only on access to current medical knowledge but also on faithfully translating changes in that knowledge into appropriate clinical recommendations. Even when updated guideline evidence is explicitly provided, a model may recognize that a recommendation has changed yet continue to retain superseded guidance, omit a critical eligibility condition, or misapply the revised recommendation. Such failures may be consequential in clinical settings, where changes in recommendation strength, treatment indications, or patient eligibility can alter appropriate care.

Many evaluations of medical LLMs, however, primarily assess performance under a fixed knowledge state. Common benchmarks examine biomedical factual knowledge, examination-style questions, and medical reasoning~\cite{jin2019pubmedqa,jin2021medqa,pal2022medmcqa,singhal2023clinical,singhal2025expert}. More recent evaluation frameworks extend beyond conventional question answering to broader medical tasks and open-ended clinical interactions~\cite{bedi2026medhelm,arora2025healthbench}. 
Studies of diagnostic reasoning, clinical decision-making, and guideline adherence have also examined the reliability and limitations of LLM-generated clinical recommendations~\cite{kanjee2023diagnostic,goh2024diagnostic,hager2024clinical,williams2024emergency,fast2024guideline,mcduff2025differential,tordjman2025deepseek}. 
Recent evaluations using script concordance testing have examined how LLMs revise diagnostic or therapeutic judgments in response to additional clinical evidence~\cite{mccoy2025clinical}. 
Related work has examined the reliability of LLMs in updating clinical judgments as longitudinal patient evidence evolves, identifying failures in how models revise their assessments in response to changing patient information~\cite{zeng2026large}. These evaluations address important aspects of clinical reasoning, including responses to changing patient-level evidence. Revisions to the underlying medical guidance introduce a complementary challenge: the recommendation itself may need to change for the same clinical situation.

Research on temporally evolving knowledge has examined time-dependent facts, temporally grounded question answering, and adaptation to new information~\cite{dhingra2022time,liska2022streamingqa,jang2022temporalwiki,mousavi2024dyknow,zeng2026medcl}.
Knowledge-editing studies have developed methods for modifying factual associations~\cite{meng2022rome,meng2023memit} and investigated whether such changes propagate consistently to related questions and downstream inferences~\cite{zhong2023mquake,cohen2024ripple,jiang2024learning}.
In medicine, TempoMed introduced longitudinal guideline trajectories and evaluated the temporal awareness of medical knowledge, documenting difficulties with historical recommendations and inconsistent responses across time, as well as limitations of agentic search in mitigating these failures~\cite{guan2026large}. FineTuneBench also evaluated the acquisition of updated knowledge, including medical guidelines, through commercial fine-tuning APIs~\cite{wu2024finetunebench}.
Together, these studies highlight the challenges of maintaining temporally appropriate medical knowledge. However, acquiring, retrieving, or recognizing updated information is distinct from faithfully incorporating its clinical consequences into a recommendation. Whether LLMs can reliably make this transition when the relevant updated guideline evidence is explicitly provided remains insufficiently understood.

Medical knowledge revisions often involve more than simple factual replacement. Recommendations may be strengthened or weakened,
eligible populations expanded or restricted, decision thresholds
modified, interventions removed, or prior guidance replaced altogether.
Correct adaptation therefore requires models to identify not only that a
change has occurred, but also what changed, which components of prior
knowledge remain valid, and how the revision should alter subsequent clinical
conclusions. 
We therefore distinguish two capabilities---revision recognition and revision
incorporation---and ask whether they dissociate in ways that conventional
evaluations of medical knowledge accuracy may fail to capture.

Here, we introduce ChronoEvo, a benchmark comprising 1,709 recommendation-level revisions across 12 forms of medical knowledge change, to systematically investigate whether LLMs faithfully incorporate evolving medical guidance into clinical recommendations. We separately assess revision recognition, clinical correctness, and full incorporation when models are provided with prior recommendations and updated guideline evidence. Across 14 models, we identify a persistent recognition–incorporation gap, in which recognition of revised guidance substantially exceeds its full incorporation into generated recommendations. Using the primary model, we further characterize revision-specific failures and examine whether prompting it to prioritize updated evidence improves incorporation. An independent semantic evaluator and adjudicated assessments by two clinicians provide complementary validation. Together, these findings identify faithful incorporation of evolving medical knowledge as a distinct challenge for clinical LLM reliability, even when updated guideline evidence is explicitly available.

\section{Results}\label{sec2}

\subsection{A longitudinal framework for evaluating evolving medical knowledge}\label{sec2.1}

\begin{figure*}[!ht]
    \centering
    \includegraphics[width=\textwidth]{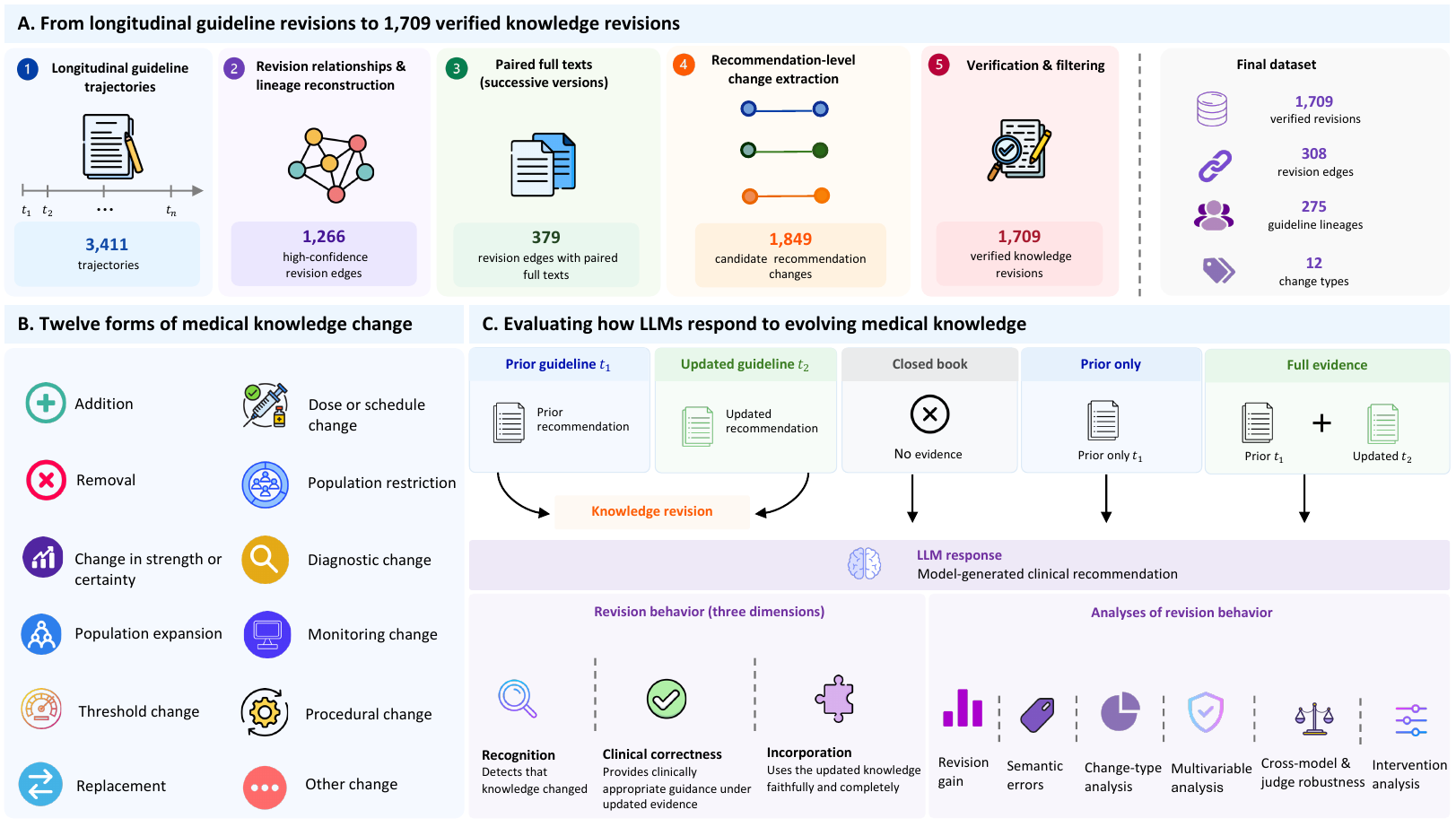}
    \vspace{0.5mm}
    \caption{\textbf{Overview of the ChronoEvo study framework.}
    a, Construction of ChronoEvo from longitudinal guideline trajectories through
    revision-relationship reconstruction, identification of revision edges with
    paired full texts, recommendation-level change extraction and verification,
    yielding 1,709 verified knowledge revisions from 308 revision edges across
    275 guideline lineages. b, Twelve semantic forms of medical knowledge change represented in the benchmark. c, Evaluation framework for testing how LLMs respond to evolving medical knowledge under closed-book, prior-only and full-evidence conditions. Model responses are evaluated along three dimensions—revision recognition, clinical correctness and full incorporation—and further examined using revision-gain, semantic-error, change-type, multivariable, cross-model and evaluator-robustness, and intervention analyses.
    Icon attributions are provided in Table S1.}
    \label{fig:overview}
\end{figure*}

To study how LLMs respond when established medical knowledge changes, we constructed ChronoEvo from longitudinal revisions of clinical practice guidelines (Fig.~\ref{fig:overview}a). ChronoEvo comprises 1,709 recommendation-level revisions retained after structured extraction and a separate model-based verification stage, spanning 308 guideline revision edges and 275 longitudinal guideline lineages. Rather than treating temporal knowledge change as a binary distinction between outdated and current information, we represented each revision as a transition from a prior recommendation to its updated recommendation, grounded in evidence from the corresponding guideline versions.

The resulting revisions captured substantial heterogeneity in how clinical knowledge evolved. We identified 12 semantically distinct forms of change, including the addition and removal of recommendations, replacement of prior guidance, changes in recommendation strength or certainty, expansions or restrictions of eligible populations, threshold changes, dose or scheduling changes, diagnostic changes, monitoring changes, and procedural changes (Fig.~\ref{fig:overview}b). Additions were the most frequent form of revision ($n=541$), followed by diagnostic changes ($n=177$), procedural changes ($n=164$), dose or scheduling changes ($n=152$), and changes in recommendation strength or certainty ($n=149$). This heterogeneity enables us to test whether LLM responses differ across forms of medical knowledge revision.

We evaluated model behavior under three controlled evidence conditions that
varied access to guideline information (Fig.~\ref{fig:overview}c). In the
full-evidence condition, models were provided with the prior recommendation
together with evidence from the updated guideline; in the prior-only
condition, the updated evidence was withheld; and in the closed-book
condition, models received no guideline evidence. This design allowed us to
quantify how model responses changed when explicit updated evidence became
available relative to conditions in which that evidence was unavailable.

\subsection{Recognition does not guarantee incorporation of revised medical knowledge}
\label{sec2.2}

In an initial evaluation using Qwen3-4B, responses generated with updated guideline evidence were more lexically aligned with the verified current recommendations than a no-update baseline that retained the prior recommendations (mean token-F1 revision gain, 0.276; 95\% CI, 0.265–0.286). Revision gain was positive for 90.2\% of revisions, indicating improvement relative to retaining the superseded recommendation.

\begin{figure*}[!ht]
    \centering
    \includegraphics[width=\textwidth]{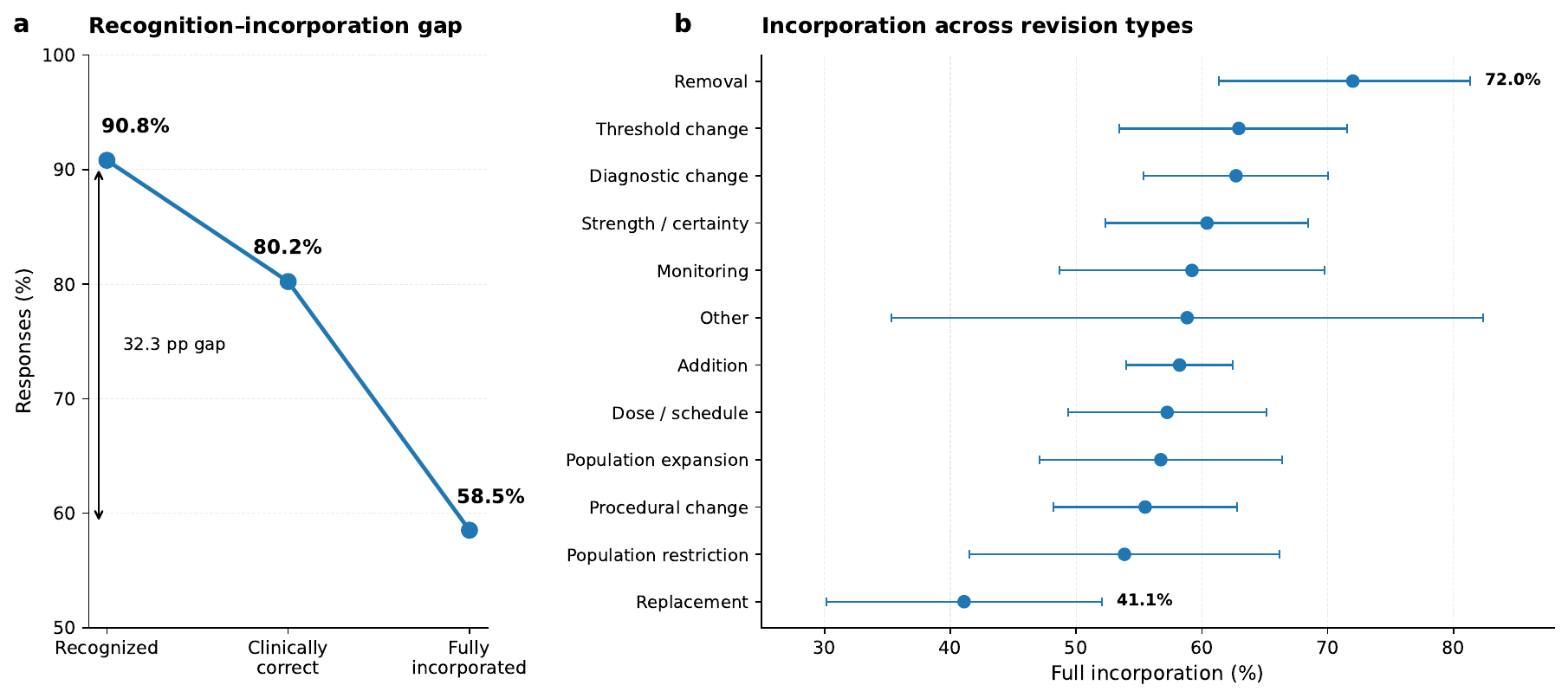}
    \caption{\textbf{Recognition and incorporation of evolving medical knowledge.}
    \textbf{a}, Proportion of responses that recognized the revision, were
    clinically correct, and fully incorporated the revised knowledge.
    \textbf{b}, Full-incorporation rates across the 12 semantic revision types.
    Error bars indicate 95\% confidence intervals. Semantic outcomes were
    evaluated by an evaluator blinded to the benchmark-assigned revision type.}
    \label{fig:revision_behavior}
\end{figure*}

We next examined whether these shifts reflected faithful semantic incorporation of the revised knowledge. Using GPT-5 as the primary semantic evaluator, blinded to the benchmark-assigned revision type, we found that 90.8\% of responses reflected recognition of the underlying revision and 80.2\% were clinically correct, whereas only 58.5\% fully incorporated the revised guidance (Fig.~\ref{fig:revision_behavior}a). This corresponds to a
32.3-percentage-point recognition--incorporation gap, indicating that
responses frequently reflected the direction or meaning of a knowledge
revision without faithfully translating all clinically relevant consequences
of that revision into the updated recommendation.

\subsection{Replacement revisions illustrate a pronounced recognition–incorporation gap}
\label{sec2.3}

The recognition--incorporation dissociation was descriptively heterogeneous
across forms of medical knowledge change
(Fig.~\ref{fig:revision_behavior}b). Full-incorporation rates ranged from
41.1\% for replacement revisions to 72.0\% for removals. This variation,
however, did not yield a statistically significant overall association
between revision type and full incorporation
($\chi^2=18.65$, $P=0.068$), and broad differences across the 12 revision
categories should therefore be interpreted cautiously.

Within this broader pattern, replacement revisions provided a particularly
clear example of recognition without complete incorporation. Although 94.5\%
of responses recognized the underlying replacement, only 41.1\% fully
incorporated the revised guidance, yielding a 53.4-percentage-point
within-category gap. In exploratory pairwise comparisons with FDR correction, replacement revisions showed a lower full-incorporation rate than removal revisions (41.1\% versus
72.0\%; FDR-adjusted $q=0.016$), the only pairwise comparison for full
incorporation that remained significant after multiple-comparison correction.
These findings suggest that recognizing that one recommendation has replaced
another does not ensure that the superseded guidance is fully displaced in
the resulting clinical recommendation.

\subsection{Outdated-content retention varies with the semantics of knowledge revision}
\label{sec2.4}

\begin{figure*}[!ht]
    \centering
    \includegraphics[width=\textwidth]{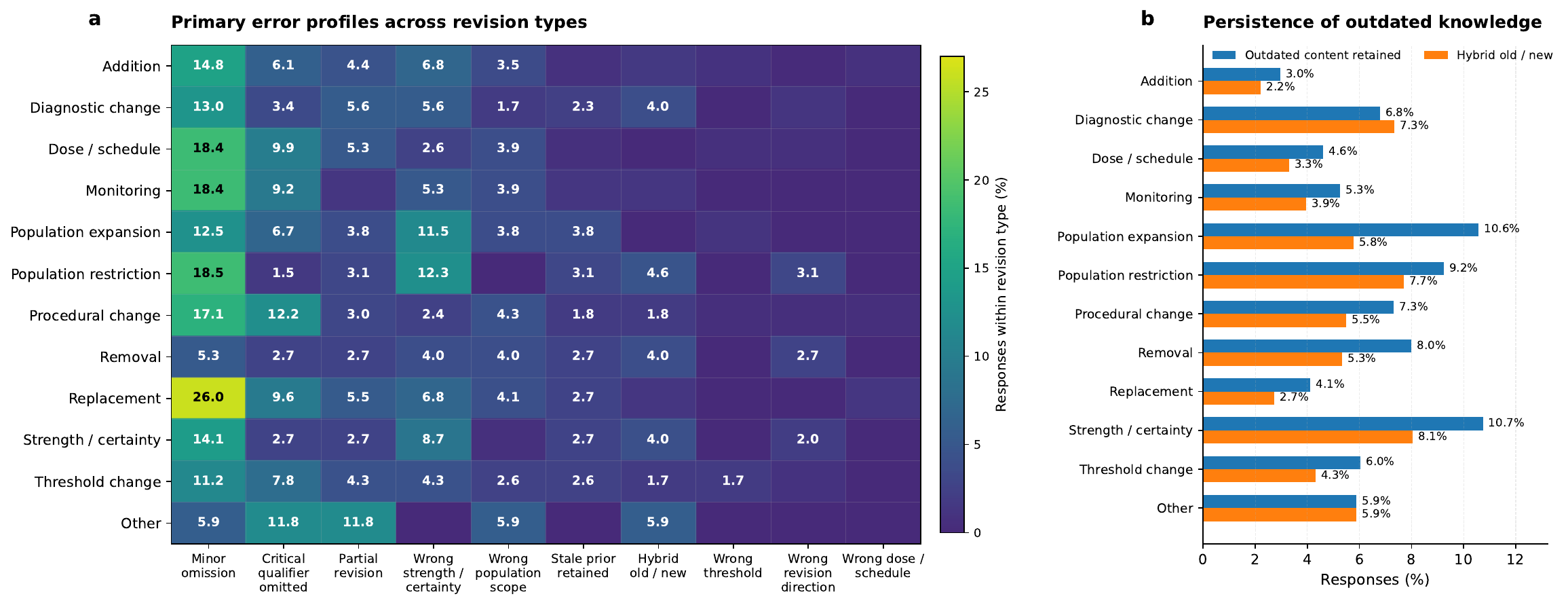}
    \caption{\textbf{Failure patterns across forms of medical knowledge evolution.}
    \textbf{a}, Distribution of primary error types across revision categories.
    Values indicate the percentage of responses within each revision type
    exhibiting the corresponding primary error.
    \textbf{b}, Two forms of persistence of superseded knowledge across revision
    categories: outdated-content retention and hybrid old--new responses.}
    \label{fig:failure_landscape}
\end{figure*}

We next asked whether incorporation failures took systematically different
forms across types of knowledge revision. Primary error profiles varied
descriptively across revision categories
(Fig.~\ref{fig:failure_landscape}a), but the overall association between
revision type and primary error category was not statistically significant
($\chi^2=166.47$, $P=0.087$; Cramer's $V=0.094$). Common error modes included minor omissions, critical qualifier omissions, incorrect recommendation strength or certainty, partial revisions, and incorrect population scope.

A more specific relationship emerged for persistence of superseded knowledge
(Fig.~\ref{fig:failure_landscape}b). Outdated-content retention differed
significantly across revision types ($\chi^2=22.46$, $P=0.021$), reaching
10.7\% for changes in recommendation strength or certainty, 10.6\% for
population expansions, and 9.2\% for population restrictions, compared with
3.0\% for additions. Pairwise analyses further showed that outdated-content
retention was significantly more frequent for strength or certainty changes
than for additions (10.7\% versus 3.0\%; FDR-adjusted $q=0.017$). Hybrid
old--new responses showed descriptive variation across revision types, but
the overall association was not statistically significant ($P=0.085$).

Thus, revision type did not strongly partition the overall landscape of
incorporation errors, but it was associated with whether superseded guidance
persisted in model responses. Revisions that modify an existing
recommendation---particularly its strength, certainty, or applicable
population---may therefore pose a different updating challenge from revisions
that introduce entirely new guidance.

\subsection{Revision-specific outdated-content retention persists after multivariable adjustment}
\label{sec2.5}

We next tested whether the revision-specific patterns persisted after
accounting for measurable differences between revision instances. We fitted
multivariable logistic-regression models for full incorporation and
outdated-content retention, adjusting for prior--current lexical overlap,
the year gap between guideline versions, and the lengths of the prior
recommendation, updated recommendation, and supporting evidence. Standard
errors were clustered by guideline revision edge to account for dependence
among recommendation-level revisions arising from the same guideline update.

\begin{figure*}[!ht]
    \centering
    \includegraphics[width=\textwidth]{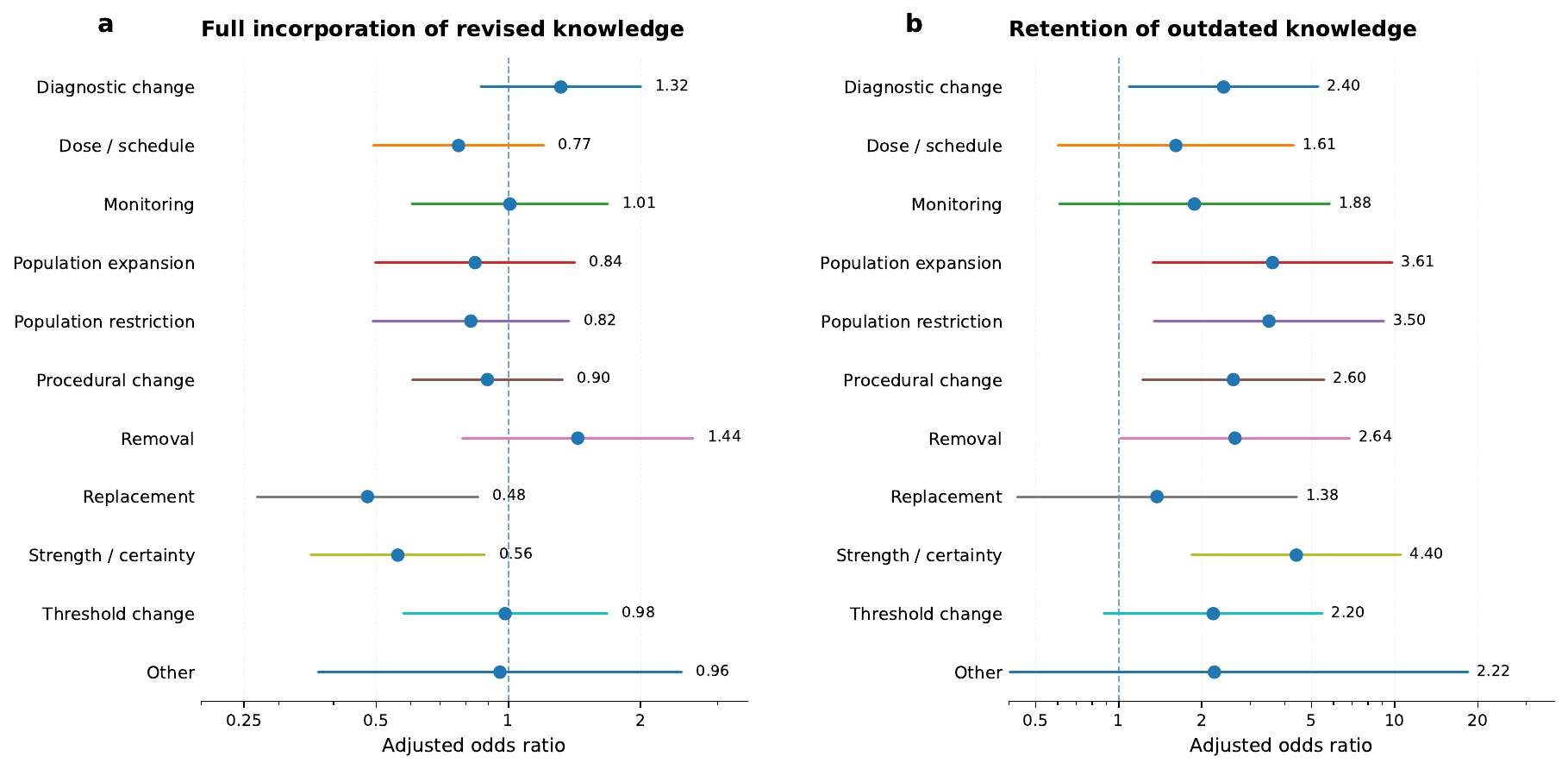}
    \caption{\textbf{Revision-type associations after adjustment for measured instance characteristics.}
    \textbf{a}, Adjusted odds ratios for full incorporation of revised knowledge
    across revision types.
    \textbf{b}, Adjusted odds ratios for retention of outdated knowledge across
    revision types.
    Addition served as the reference category. Models were adjusted for
    prior--current lexical overlap, year gap between guideline versions, and
    the lengths of the prior recommendation, updated recommendation, and
    supporting evidence. Points indicate adjusted odds ratios and horizontal
    lines indicate 95\% confidence intervals; dashed vertical lines denote an
    odds ratio of 1. Standard errors were clustered by guideline revision edge.}
    \label{fig:multivariable_forest}
\end{figure*}

For full incorporation, replacement revisions had lower adjusted odds than
additions (adjusted OR, 0.48; 95\% CI, 0.27--0.85; $P=0.013$), as did
changes in recommendation strength or certainty (adjusted OR, 0.56; 95\% CI,
0.36--0.88; $P=0.013$) (Fig.~\ref{fig:multivariable_forest}a). Neither
association remained significant after Benjamini--Hochberg correction across
the 11 revision-type coefficients (both FDR-adjusted $q=0.070$). 
In contrast, several instance-level characteristics showed associations with full incorporation. Greater prior--current lexical similarity was associated with higher odds of full incorporation (adjusted OR per s.d. increase, 1.65; 95\% CI, 1.47--1.86), whereas longer updated recommendations were associated with lower odds (adjusted OR per s.d. increase, 0.47; 95\% CI, 0.38--0.57). Longer prior recommendations were associated with modestly higher odds of incorporation (adjusted OR per s.d. increase, 1.17; 95\% CI, 1.03--1.32), whereas greater temporal separation between guideline versions was associated with slightly lower odds (adjusted OR per s.d. increase, 0.90; 95\% CI, 0.81--1.00).

A different pattern emerged for outdated-content retention
(Fig.~\ref{fig:multivariable_forest}b). Relative to additions, changes in
recommendation strength or certainty were associated with substantially
higher odds of retaining outdated content (adjusted OR, 4.40; 95\% CI,
1.84--10.54; FDR-adjusted $q=0.010$). Population expansions (adjusted OR,
3.61; 95\% CI, 1.33--9.80; $q=0.037$), population restrictions (adjusted OR,
3.50; 95\% CI, 1.34--9.14; $q=0.037$), and procedural changes (adjusted OR,
2.60; 95\% CI, 1.22--5.55; $q=0.037$) were likewise associated with greater
outdated-content retention after multiple-comparison correction. None of the
measured continuous instance-level characteristics was significantly
associated with outdated-content retention after multivariable adjustment.

These associations were insensitive to the level at which dependence was
modeled. When standard errors were instead clustered by longitudinal
guideline lineage, the same four revision categories remained significantly
associated with outdated-content retention after FDR correction. Together, these analyses suggest different patterns of association for the two outcomes: full incorporation was associated with several
instance-level characteristics, whereas outdated-content retention
showed significant associations with specific revision types after
multiple-comparison correction.

\subsection{The recognition--incorporation gap persists across LLMs}
\label{sec2.6}

The recognition--incorporation gap was not specific to the primary model.
Across all 14 evaluated models, spanning multiple model families, parameter
scales, medical specialization, and capability tiers, revision recognition
consistently exceeded full incorporation of the revised knowledge
(Table~\ref{tab:cross_model}). Among open-weight models, recognition ranged
from 87.4\% to 94.1\%, whereas full incorporation ranged from 50.7\% to
67.5\%, leaving gaps of 26.4--37.4 percentage points. Cluster-bootstrap
confidence intervals for the gap excluded zero for every evaluated model
(Table S3)
, indicating that the dissociation generalized
beyond the primary experimental setting.

Increasing model scale improved absolute performance in several families but
did not reliably close this gap. Scaling Qwen3 from 4B to 32B increased
recognition by 3.2 percentage points (95\% CI, 1.7 to 4.7), clinical
correctness by 2.8 points (0.6 to 4.9), and full incorporation by 4.7 points
(2.2 to 7.2), yet changed the gap by only -1.5 points (-3.9 to 0.8).
Similarly, scaling Llama-3.1 from 8B to 70B improved all three absolute
endpoints while changing the gap by -2.2 points (-4.6 to 0.1), and scaling
MedGemma from 4B to 27B changed the gap by -1.4 points (-3.9 to 1.1).
For Mistral, the larger model improved recognition and clinical correctness
without improving full incorporation, and the gap increased numerically.
None of the paired within-family scale comparisons produced a statistically
significant reduction in the recognition--incorporation gap
(Table S4).

Greater frontier-model capability likewise reduced, but did not eliminate,
the absolute dissociation. GPT-5.6 Luna, Terra, and Sol achieved
full-incorporation rates of 75.9\%, 76.5\%, and 77.9\%, respectively,
compared with recognition rates of 97.2\%, 97.1\%, and 98.0\%. GPT-5.6 Sol,
the strongest model evaluated, therefore retained a
20.1-percentage-point gap (95\% CI, 17.9 to 22.4). Relative to Luna, Sol
significantly improved clinical correctness and full incorporation, but the
change in the gap was not statistically significant (-1.2 percentage points;
95\% CI, -2.8 to 0.3). Improvements in overall model capability therefore
translated into better updated recommendations without proportionately
eliminating the distinction between recognizing and fully incorporating a
knowledge revision.

Finally, the central finding was robust to the choice of semantic evaluator.
Independent re-evaluation of all 1,709 primary-model predictions with
GPT-5.6 Sol yielded raw agreement of 94.5\% for revision recognition, 90.0\%
for clinical correctness, and 88.5\% for full incorporation, with Cohen's
$\kappa$ values of 0.566, 0.699, and 0.758, respectively
(Table S2)
. Although the evaluators differed in their
absolute endpoint rates, they produced nearly identical
recognition--incorporation gaps: 32.3 percentage points with GPT-5 and
31.4 percentage points with GPT-5.6 Sol. The observed dissociation therefore
did not depend on a particular model family, scale, capability tier, or
semantic evaluator.

\begin{table*}[!ht]
\centering
\caption{\textbf{The recognition--incorporation gap persists across model scales and families.}
All models were evaluated on the same 1,709 knowledge revisions using a semantic evaluator blinded to the benchmark-assigned revision type.
Qwen3-4B, shown in bold, is the primary backbone used in the main experiments.
Clinical denotes clinical correctness; gap denotes the percentage-point difference between revision recognition and full incorporation.}
\label{tab:cross_model}
\renewcommand{\arraystretch}{1.15}
\begin{adjustbox}{max width=0.96\textwidth}
\begin{tabular}{lcccc}
\toprule
\textbf{Model} &
\textbf{Recognition (\%)} &
\textbf{Clinical (\%)} &
\textbf{Full incorporation (\%)} &
\textbf{Gap (pp)} \\
\midrule

\textbf{Qwen3-4B}
& \textbf{90.8}
& \textbf{80.2}
& \textbf{58.5}
& \textbf{32.3} \\

Qwen3-14B
& 91.3
& 82.4
& 62.3
& 29.0 \\

Qwen3-32B
& 94.0
& 83.0
& 63.2
& 30.8 \\

\midrule

Mistral-7B-Instruct-v0.3
& 89.8
& 78.0
& 58.2
& 31.6 \\

Mistral-Small-24B-Instruct-2501
& 91.9
& 80.7
& 57.7
& 34.2 \\

\midrule

Phi-4
& 93.9
& 84.6
& 67.5
& 26.4 \\

\midrule

Gemma-3-12B-IT
& 92.5
& 86.0
& 55.1
& 37.4 \\

\midrule

MedGemma-4B-IT
& 88.9
& 81.9
& 53.8
& 35.1 \\

MedGemma-27B-Text-IT
& 94.1
& 82.6
& 60.4
& 33.7 \\

\midrule

Llama-3.1-8B-Instruct
& 87.4
& 74.0
& 50.7
& 36.7 \\

Llama-3.1-70B-Instruct
& 94.1
& 84.1
& 59.6
& 34.5 \\

\midrule

GPT-5.6 Luna
& 97.2
& 89.2
& 75.9
& 21.3 \\

GPT-5.6 Terra
& 97.1
& 91.0
& 76.5
& 20.6 \\

GPT-5.6 Sol
& 98.0
& 91.6
& 77.9
& 20.1 \\

\bottomrule
\end{tabular}
\end{adjustbox}
\end{table*}

\subsection{Revision salience reduces outdated-content errors but does not close the incorporation gap}
\label{sec2.7}

We next tested a simple explanation for incomplete incorporation: that models
recognized the revision but failed to give the updated evidence sufficient
priority over the prior recommendation. We therefore introduced a
revision-salience intervention that explicitly instructed the model to
prioritize updated guideline evidence over conflicting or outdated prior
guidance (Table~\ref{tab:revision_salience}).

Increasing revision salience improved several aspects of model behavior.
Clinical correctness increased from 80.2\% to 82.6\%
($\Delta=+2.4$ percentage points, 95\% cluster-bootstrap CI, +0.7 to +4.0;
$P=0.005$), while outdated-content retention decreased from 5.9\% to 4.3\%
($\Delta=-1.6$ pp, 95\% CI, -2.6 to -0.6; $P=0.001$). Hybrid old--new
responses likewise decreased from 4.5\% to 2.7\% ($\Delta=-1.8$ pp,
95\% CI, -2.7 to -0.9; $P<0.0001$). Explicitly prioritizing the updated
evidence therefore reduced overt carryover and mixing of superseded guidance.

These improvements were not accompanied by a statistically detectable increase in full incorporation. Recognition changed only modestly from 90.8\%
to 91.7\% ($\Delta=+0.9$ pp, 95\% CI, -0.4 to +2.3), while full
incorporation remained essentially unchanged at 58.5\% versus 59.3\%
($\Delta=+0.8$ pp, 95\% CI, -1.1 to +2.7). Consequently, the
recognition--incorporation gap was virtually identical under standard and
revision-salience prompting (32.3 versus 32.4 percentage points;
$\Delta=+0.1$ pp, 95\% CI, -2.0 to +2.2).

The intervention therefore separates suppression of overt outdated content
from faithful incorporation of revised knowledge. Making the updated evidence
more salient reduced residual old guidance and improved clinical correctness,
but was insufficient to ensure that all clinically relevant consequences of
the revision were propagated into the final recommendation. These findings suggest that insufficient prioritization of updated evidence alone does not fully account for the observed recognition--incorporation gap, although other approaches to evidence prioritization remain to be tested.

\begin{table}[t]
\centering
\caption{\textbf{Effect of revision-salience prompting on knowledge updating.}
Differences are paired percentage-point changes relative to the standard
full-evidence condition. Confidence intervals were obtained using 10,000
bootstrap resamples clustered by guideline revision edge. Differences were calculated from unrounded case-level estimates and may
therefore differ by 0.1 percentage points from differences obtained from
the displayed rounded values.}
\label{tab:revision_salience}
\small
\begin{tabular}{lccc}
\toprule
\textbf{Outcome} &
\textbf{Standard (\%)} &
\textbf{Revision salience (\%)} &
\textbf{$\Delta$ (pp; 95\% CI)} \\
\midrule
Revision recognition
& 90.8 & 91.7 & +0.9 [-0.4, +2.3] \\

Clinical correctness
& 80.2 & 82.6 & +2.4 [+0.7, +4.0] \\

Full incorporation
& 58.5 & 59.3 & +0.8 [-1.1, +2.7] \\

Outdated retention
& 5.9 & 4.3 & -1.6 [-2.6, -0.6] \\

Hybrid old/new
& 4.5 & 2.7 & -1.8 [-2.7, -0.9] \\

\midrule
Recognition--incorporation gap
& 32.3 & 32.4 & +0.1 [-2.0, +2.2] \\
\bottomrule
\end{tabular}
\end{table}

\subsection{Clinician assessment corroborates the recognition--incorporation gap}
\label{sec2.8}
To assess whether the recognition--incorporation gap identified by automated semantic evaluation reflected clinically meaningful differences rather than evaluator-specific artifacts, we conducted a clinical assessment of 300 model responses comprising 250 cases from the primary analysis set and 50 cases from a targeted enrichment set. Two clinician annotators, blinded to the automated labels, independently evaluated whether each response recognized the underlying knowledge revision, provided a clinically correct recommendation, and fully incorporated the updated guidance. Disagreements were resolved by consensus discussion to establish an adjudicated reference assessment.

Clinician evaluation reproduced the central recognition--incorporation gap. Of the 300 responses, 277 (92.3\%) were judged to recognize the underlying revision and 238 (79.3\%) were clinically correct, whereas only 156 (52.0\%) fully incorporated the updated recommendation, corresponding to a 40.3-percentage-point gap between recognition and full incorporation. Evaluation of the same responses by the automated semantic judge (GPT-5) yielded recognition and full incorporation rates of 91.3\% and 57.3\%, respectively, corresponding to a 34.0-percentage-point gap. Thus, the discrepancy between recognizing revised medical knowledge and fully incorporating it persisted under clinician assessment. Moreover, among the 144 responses judged not to have fully incorporated the revision, 61 (42.4\%) were judged clinically incorrect, demonstrating that a substantial subset of incomplete incorporations involved clinically substantive errors.

Inter-annotator agreement was 96.3\%, 86.0\%, and 88.0\% for revision recognition, clinical correctness, and full incorporation, respectively, with Cohen's $\kappa$ values of 0.683, 0.644, and 0.760; the comparatively lower $\kappa$ for revision recognition, despite high raw agreement, is consistent with the high prevalence of positive labels. Agreement between GPT-5 and the adjudicated clinician assessments was 99.0\%, 89.7\%, and 85.3\% ($\kappa$ = 0.933, 0.698, and 0.705), indicating substantial concordance with clinician assessments and providing empirical support for the use of automated semantic evaluation to characterize revision behavior at scale, while highlighting residual disagreement in clinically nuanced judgments. Within the validation sample, GPT-5 classified a higher proportion of responses as fully incorporating the revision than clinicians (57.3\% versus 52.0\%).

Together, these findings provide clinician-based corroboration of the recognition--incorporation gap observed in the large-scale automated evaluation. Although models frequently reflected updated medical knowledge, they substantially less often incorporated the revised recommendations in full, indicating that recognition alone is insufficient to establish reliable clinical knowledge updating.

\section{Discussion}
\label{sec:discussion}

Our findings reveal a systematic dissociation between recognizing revised medical knowledge in a generated response and faithfully incorporating its consequences into an updated clinical recommendation. Across 1,709 recommendation-level revisions, the primary model
achieved 90.8\% revision recognition but only 58.5\% full
incorporation, yielding a 32.3-percentage-point gap. The dissociation persisted across 14 models and two semantic evaluators and was further corroborated by an adjudicated assessment of 300 responses by two clinicians. A targeted revision-salience intervention reduced outdated-content errors but did not produce a statistically detectable improvement in full incorporation. Together, these results identify faithful incorporation of revised guidance as a distinct dimension of observable clinical LLM behavior.

This distinction extends existing concerns about temporal knowledge in LLMs.
Prior work on knowledge currency has established that models may retain
outdated information or fail to identify the most current medical knowledge.
ChronoEvo addresses a complementary downstream problem: even when updated
guideline evidence is explicitly available, models may fail to translate that
evidence faithfully into the recommendation that follows from it. The
full-evidence condition generally shifted responses toward current guidance,
with positive revision gain for 90.2\% of revisions, indicating that models
were responsive to the updated evidence. Nevertheless, semantic evaluation
revealed substantially lower rates of full incorporation. Access to current evidence therefore does not, by itself, guarantee reliable knowledge updating. Evaluations that assess only
whether a model can identify or retrieve current information may consequently
overestimate its ability to revise downstream recommendations when the
underlying knowledge changes.

The structure of these failures provides further insight into the challenge.
Although broad differences in full incorporation and primary error profiles
across all revision categories did not reach conventional statistical
significance, specific patterns were evident. 
Replacement revisions exhibited a particularly pronounced
descriptive recognition--incorporation gap, whereas
outdated-content retention varied significantly across
revision types.
Changes in recommendation strength or certainty, population scope, and
procedures remained associated with greater outdated-content retention after
multivariable adjustment and multiple-comparison correction. These findings
suggest that knowledge updating is not a homogeneous operation: different
forms of revision may impose different requirements on how prior and current
information must be reconciled, with some revisions being particularly prone
to residual persistence of superseded guidance.

The persistence of the recognition--incorporation gap across models also has
implications for model scaling and domain specialization. Larger models generally achieved higher absolute performance,
and the evaluated frontier models exhibited smaller
recognition--incorporation gaps than most open-weight models.
Nevertheless, the gap remained substantial even in the
strongest model evaluated, and within-family scaling did not
produce a statistically significant reduction in the gap.
Medically specialized models also exhibited substantial
recognition--incorporation gaps. These findings do not
imply that scale or domain-specific training is unhelpful; both may improve
overall clinical performance. Rather, they indicate that improvements in
general capability or medical knowledge do not necessarily produce
proportionate improvements in faithful knowledge updating. This distinction
may become increasingly important as benchmark performance approaches ceiling
levels on tasks that primarily measure factual knowledge or recognition.

The revision-salience intervention helps clarify the limits of a simple prompting-based approach to knowledge updating. Explicitly instructing the model to prioritize updated guideline evidence improved clinical correctness and reduced outdated-content retention and hybrid old--new responses. However, the intervention did not yield a statistically detectable improvement in full incorporation or a reduction in the recognition--incorporation gap. These findings suggest that emphasizing the priority of revised evidence, at least through the prompting strategy tested here, is insufficient to ensure that all clinically relevant consequences of a guideline revision are reflected in the final recommendation.

These findings have practical implications for the evaluation and deployment
of clinical LLMs. In settings where guidelines evolve, a response can appear
current and clinically plausible while retaining an outdated qualifier,
population restriction, recommendation strength, or component of a
superseded treatment strategy. Incomplete incorporation does not necessarily imply clinical incorrectness, because a response may remain clinically consistent with updated guidance while omitting some components of the revision. Nevertheless, in the clinician-assessed sample, 61 of 144 responses (42.4\%) judged to have incompletely incorporated the revision were also judged clinically incorrect. This distinction underscores the importance of evaluating both clinical correctness and completeness of revision incorporation. Evaluation of clinical LLMs under evolving knowledge should therefore
distinguish whether models recognize that guidance has changed from whether
they faithfully incorporate the consequences of that change. Revision-aware evaluation may be particularly important for
revisions with potentially consequential clinical implications,
including changes in treatment eligibility, recommendation
strength, or the replacement or withdrawal of prior guidance.
In such cases, incomplete updating could alter the clinical
meaning of an otherwise plausible recommendation.

Several limitations should be considered. First, ChronoEvo evaluates
knowledge evolution through revisions to clinical practice guidelines.
Guidelines provide an unusually well-defined setting in which prior and
updated recommendations can be compared longitudinally, but they represent
only one pathway through which medical knowledge changes. Future work should
extend this framework to drug safety communications, regulatory actions,
emerging trial evidence, consensus statements, and other sources of evolving
medical knowledge. Second, our evaluation focuses on model outputs and does
not establish the internal mechanisms responsible for the observed
recognition--incorporation gap. Revision recognition is operationalized
behaviorally from generated responses rather than as a direct measurement of
a model's internal knowledge state. Accordingly, the present findings
demonstrate a dissociation in observable behavior but do not imply a specific
mechanistic separation within the model. Third, although we evaluated diverse
open-weight, medically specialized, and proprietary models, the model
landscape continues to evolve, and the magnitude of the gap may change with
future architectures and training strategies. 
Fourth, benchmark construction and large-scale semantic evaluation relied partly on automated model judgments. GPT-5 was used for recommendation extraction, verification, and primary response evaluation, creating potential shared model-dependent biases despite the separation of these stages. Independent re-evaluation with GPT-5.6 Sol and clinician assessment of 300 responses provide complementary evidence for the central finding, but do not constitute exhaustive expert verification of the benchmark revisions. The clinician sample also included targeted enrichment of selected revision types and was assessed by two physicians from different surgical specialties; its absolute endpoint rates should therefore be interpreted as descriptive of the validation sample rather than generalized to the full benchmark or all clinical specialties.
Finally, ChronoEvo evaluates whether models can
incorporate revised knowledge when relevant guideline evidence is provided;
it does not test the separate challenges of retrieving the correct guideline,
determining which version applies to a particular patient or time point, or
maintaining updated knowledge over prolonged real-world deployment. How these additional challenges interact with the incorporation failures identified here remains an
important question for future research.

More broadly, our results suggest that keeping an LLM clinically current
requires more than exposing it to current information. A model must determine
how new evidence changes prior guidance, preserve components that remain
valid, remove those that have been superseded, and propagate the revision
faithfully into the recommendation it produces. ChronoEvo provides a framework
for evaluating this process directly. The persistent gap between revision
recognition and full incorporation identifies faithful knowledge incorporation
as a distinct challenge for clinical LLMs and motivates evaluation and
adaptation methods designed not merely to recognize evolving medical
knowledge, but to update the recommendations that depend on it.

\section{Methods}\label{sec:methods}

\subsection{Construction of longitudinal medical knowledge revisions}

We constructed a collection of naturally occurring medical knowledge revisions
from longitudinal versions of clinical practice guidelines. We started from
the 3,411 guideline trajectories in TempoMed-Traj~\cite{guan2026large}, which
organize historical and current guideline documents addressing related
clinical topics. Because document-level temporal relations in these
trajectories do not necessarily correspond to direct successive revisions of
the same guideline, we first reconstructed a conservative graph of guideline
evolution before extracting recommendation-level changes.

For each TempoMed-Traj trajectory, historical--current guideline relations
were treated as candidate revision links and enriched with bibliographic
metadata. Candidate links were validated using GPT-5 (temperature = 1) with
structured output. The validator classified each document pair as a direct
revision, probable revision, same-version parallel publication, summary or
companion publication, related guideline, commentary or review, or uncertain
relationship. Validation considered the clinical topic, publication
chronology, document titles, issuing organizations, journals, publication
types, PubMed identifiers, the original TempoMed relationship rationale, and
pair-level similarity signals. The validation prompt was deliberately
precision-oriented: chronological ordering, topic or title similarity,
organizational overlap, or citation of an earlier guideline was not by itself
considered sufficient evidence of a direct revision. A direct-revision label
required positive evidence that the later document updated, revised, replaced,
superseded, or otherwise succeeded the earlier guideline.

To construct the high-precision revision graph, we retained only candidate
relations classified as direct revisions for which the validator additionally
confirmed membership in the same guideline lineage and an explicit revision
signal. Probable revisions and relations classified as parallel publications,
summaries or companion documents, related but distinct guidelines,
commentaries or reviews, or uncertain were excluded from the primary graph.
Publication-level records were then canonicalized to reduce duplicate
manifestations of the same guideline version. Within the same clinical-topic
context, records from the same year whose normalized titles had a token
Jaccard similarity of at least 0.82 were grouped as candidate parallel or
co-publication manifestations and collapsed into a single canonical
guideline-version node. Publication-level revision relations were subsequently
collapsed onto these canonical nodes, and relations whose source and target
collapsed to the same canonical version were removed.

We next reconstructed longitudinal guideline families from the canonical
revision graph and searched for potentially missing revision relations between
chronologically adjacent canonical versions. Each candidate missing link
underwent a second structured GPT-5 validation (temperature = 1) using the
complete chronological family context, already established adjacent
relationships, and pair-level signals including year gap, title similarity,
organizational overlap, and shared clinical topics. Candidate links were
classified as direct revision, probable revision, not revision, or uncertain.
Chronological adjacency or metadata similarity alone was explicitly considered
insufficient evidence of revision. Direct revisions recovered at this stage
were incorporated into the primary Core graph. Probable revisions were
excluded from the Core graph and retained only in a separate Extended graph
when validation confidence was at least 0.75 and both same-guideline lineage
and temporal consistency were confirmed.

We used the Core graph for construction of the primary ChronoEvo benchmark.
Before deriving final lineages, edges inconsistent with forward chronology
were removed, any residual graph cycles were resolved by removing the
lowest-confidence edge in the cycle, and transitive reduction was performed
such that a direct edge was removed only when an alternative graph path
already connected the same source and target. This procedure produced a
high-confidence graph containing 1,266 guideline-level revision edges.

We next identified 379 revision edges for which usable full text was available
for both the earlier and updated guideline versions. For each paired edge,
guideline documents in XML/NXML, HTML, or plain-text format were converted to
plain text and whitespace-normalized, with each document limited to the first
120,000 characters. The earlier and updated guideline texts, together with
their publication years and titles, were provided to GPT-5
(temperature = 1) using a structured extraction prompt. The model was
instructed to identify clinically substantive changes in recommendations
rather than differences limited to wording, formatting, organization,
citations, background discussion, literature review, methodology, or evidence
summaries, and to favor precision over recall. Both the prior and updated
recommendations were required to be supported by their respective source
documents, and increased detail in the later guideline alone was not treated
as evidence of a clinical change.

Extraction used a predefined structured-output schema. For each candidate
revision, GPT-5 returned the clinical focus, prior recommendation, updated
recommendation, semantic change type, a concise change summary, supporting
evidence spans from the earlier and updated guidelines, and an indicator of
whether the difference represented a substantive clinical change. The semantic
change type was constrained to one of 12 predefined categories: addition,
removal, replacement, strength or certainty change, population expansion,
population restriction, threshold change, dose or schedule change, diagnostic
change, monitoring change, procedural change, or other. A single guideline
revision edge could yield multiple independent recommendation-level changes;
when no reliable substantive change could be established, an empty set was
returned. This procedure yielded 1,849 candidate recommendation-level
revisions.

Candidate revisions were subsequently subjected to a separate structured
GPT-5 verification stage (temperature = 1). For each candidate, the verifier
was provided with the candidate revision and the earlier and updated guideline
texts and was instructed to independently reassess the candidate rather than
trust the extraction output or its supporting evidence spans. The verifier
separately determined whether (1) the earlier guideline supported the stated
prior recommendation, (2) the later guideline supported the stated updated
recommendation, (3) the two recommendations concerned the same or directly
comparable clinical focus, (4) the recommendation meaningfully changed across
guideline versions, and (5) the difference constituted a clinically
substantive change rather than stylistic rewriting, clarification,
reorganization, or a background or evidence-summary update.

A candidate was retained only when all five verification criteria were
satisfied and the verification confidence was at least 0.80. The verifier
additionally assessed whether the semantic change type assigned during
extraction appropriately characterized the verified revision. When the
original category was judged incorrect and a corrected category was provided,
the corrected category was used without otherwise excluding the candidate.
Of the 1,849 extracted candidates, 1,709 satisfied these verification
criteria.

We additionally applied conservative within-edge duplicate detection to the
verified revisions. Two candidates were considered duplicates only when their
clinical focus, prior recommendation, and updated recommendation each had a
token-level Jaccard similarity of at least 0.85. Candidates were processed in
descending order of verification confidence so that higher-confidence
candidates were preferentially retained. No additional duplicates met these
criteria. Of the 379 guideline revision edges with paired full texts, 308
contributed at least one retained recommendation-level
revision. The final ChronoEvo benchmark therefore contained
1,709 verified recommendation-level revisions derived from
these 308 edges and 275 longitudinal guideline lineages.

Each verified revision was represented by its clinical focus, prior
recommendation, updated recommendation, supporting evidence from the
corresponding guideline versions, publication years, and semantic change type.
The final collection comprised 541 additions, 75 removals, 73 replacements,
149 strength or certainty changes, 104 population expansions, 65 population
restrictions, 116 threshold changes, 152 dose or schedule changes, 177
diagnostic changes, 76 monitoring changes, 164 procedural changes, and 17
revisions categorized as other. Each recommendation-level revision contained
a single semantic change-type field in the structured representation.
Semantic change type was retained as benchmark metadata for stratified
analyses and was not provided to the generation models or to the semantic
evaluator.

For dataset partitioning, we used lineage-level rather than sample-level
splitting to prevent leakage across related guideline versions. All revisions
belonging to the same reconstructed lineage group were restricted to a single
partition, and lineage groups were allocated to training, development, and
test sets with target proportions of 80\%, 10\%, and 10\%, respectively,
while approximately balancing the number of recommendation-level revisions.
This yielded 1,367 training, 171 development, and 171 test revisions, with no
lineage group, guideline revision edge, or guideline version shared across
partitions. These partitions are provided as leakage-safe benchmark splits for
future model-development settings; the zero-shot evaluations reported in this
study used all 1,709 revisions because no ChronoEvo samples were used for
model training or adaptation.


\subsection{Evaluation tasks and evidence conditions}

We evaluated whether LLMs could update clinical recommendations under
different levels of access to longitudinal guideline information. Each
evaluation instance was defined by a clinical focus and a verified transition
from a prior to an updated recommendation. The updated recommendation served
exclusively as the reference target for evaluation and was never included in
the generation prompt. Models were not provided with the semantic change type
or other benchmark annotations.

We evaluated three evidence conditions. In the \textit{full-evidence}
condition, the model received the clinical focus, the earlier recommendation
with its publication year, and supporting evidence from the updated guideline
with the corresponding publication year. The model was explicitly asked to
update the earlier recommendation on the basis of the newly available
guideline evidence and to provide the resulting recommendation directly and
concisely. This condition therefore tested whether a model could translate
explicit evidence of a guideline revision into an updated recommendation.

In the \textit{prior-only} condition, the model received the clinical focus
and the earlier recommendation with its publication year, but no evidence
from the updated guideline or information about the updated guideline year.
It was asked to provide the recommendation according to current clinical
guidance. This condition assessed the recommendation produced from the
model's existing knowledge when anchored by the earlier recommendation but
without access to the evidence supporting its revision.

In the \textit{closed-book} condition, the model received only the clinical
focus and was asked to provide the recommendation according to current
clinical guidance. Neither the earlier recommendation, guideline evidence,
nor guideline-version years were provided. This condition assessed the
recommendation available from the model's existing knowledge without
longitudinal guideline context.

The full-evidence condition served as the primary setting for subsequent
analyses of revision recognition, clinical correctness, full incorporation,
failure modes, and cross-model robustness. The complete system and user
prompts for all three evidence conditions are provided in 
Supplementary Note 2.

\subsection{Model generation and experimental settings}

Qwen3-4B was designated as the primary model for detailed analyses of
knowledge updating. The same set of 1,709 verified recommendation-level
revisions was evaluated under each of the three evidence conditions. Model
outputs were generated independently for each revision without access to the
reference updated recommendation, semantic change type, or evaluation labels.

Qwen3-4B was evaluated using the Hugging Face Transformers library with its
instruction-tuned chat template. Each input was formatted as a two-message
conversation comprising a condition-specific system instruction and the
corresponding user prompt. Qwen3's thinking mode was disabled
(\texttt{enable\_thinking=False}). Each benchmark instance was processed in
an independent conversational context, such that information from other
guideline revisions could not enter the model context.

Inputs were tokenized with left padding and right truncation to a maximum
input length of 2,048 tokens. Responses were generated using deterministic
greedy decoding (\texttt{do\_sample=False}) with a maximum of 256 new tokens.
Generation used the model-specific end-of-sequence token for termination and
cached key--value states during autoregressive decoding. Samples were
processed in batches of four. No sampling temperature, top-$p$, or other
stochastic decoding parameters were used.

For all evidence conditions, the reference updated recommendation was used
only for post-generation evaluation and was never included in the model
prompt. The semantic change type and all evaluation labels were likewise
withheld from the generation model. Condition-specific inputs were constructed
as described above, and the complete system and user prompts are provided in
Supplementary Note 2.

\subsection{Evaluation of knowledge revision and incorporation}

We evaluated model responses using complementary lexical and semantic
measures. To quantify whether access to updated guideline evidence moved a
generated recommendation away from superseded guidance and toward the
reference updated recommendation, we defined a per-instance revision gain.
Let $\hat{y}^{\,\mathrm{full}}_i$ denote the recommendation generated under
the full-evidence condition, $y^{\,\mathrm{prior}}_i$ the earlier guideline
recommendation, and $y^{\,\mathrm{current}}_i$ the verified updated
recommendation. Revision gain was defined as

\begin{equation}
G_i =
\mathrm{F1}\!\left(
\hat{y}^{\,\mathrm{full}}_i,
y^{\,\mathrm{current}}_i
\right)
-
\mathrm{F1}\!\left(
y^{\,\mathrm{prior}}_i,
y^{\,\mathrm{current}}_i
\right),
\end{equation}

where F1 denotes token-level F1 after lowercasing, whitespace normalization,
and removal of punctuation. The second term represents a natural no-update
baseline in which the earlier recommendation is retained as the answer to the
current-recommendation question. Positive revision gain therefore indicates
that the response generated with updated guideline evidence was lexically
closer to the verified current recommendation than the superseded
recommendation itself. A value of zero indicates no improvement over retaining
the earlier recommendation, whereas a negative value indicates movement away
from the current recommendation relative to this baseline.

Because lexical similarity does not establish clinical fidelity, we also
performed structured semantic evaluation of each full-evidence response. A
GPT-5 evaluator received the clinical focus, the earlier recommendation, the
updated guideline evidence, the verified current recommendation, and the
model-generated recommendation. The evaluator was explicitly blinded to the
benchmark-assigned semantic revision type and did not receive any downstream
analysis labels. Evaluation focused on clinically meaningful changes rather
than lexical overlap or stylistic similarity.

For each response, the evaluator produced five binary semantic outcomes:
whether the generated response reflected the substantive direction or meaning of the guideline revision (revision recognition); whether the response was clinically consistent
with the verified current recommendation (\textit{clinically correct});
whether all clinically important components of the revision were faithfully
incorporated (\textit{fully incorporates revision}); whether clinically
meaningful superseded guidance was incorrectly retained
(\textit{preserves outdated content}); and whether incompatible elements of
the earlier and updated recommendations were combined
(\textit{mixes old and new}). The evaluator additionally assigned a single
primary error category, an error-severity level, the direction of the revision,
a brief explanation, and an evaluation confidence score.

Primary error categories comprised fully correct, minor omission, partial
revision, stale prior retained, hybrid old/new, wrong revision direction,
critical qualifier omitted, unsupported addition, wrong replacement, wrong
population scope, wrong threshold, wrong dose or schedule, wrong strength or
certainty, and other substantive error. The evaluator was instructed to ignore
harmless paraphrases and to judge clinical meaning, including changes in
population scope, direction, recommendation strength, thresholds, treatment
choice, dose or schedule, monitoring, procedures, and diagnostic implications.
The verified current recommendation served as the reference state, with the
earlier recommendation used to determine what had changed and the updated
guideline evidence used as supporting context.

Semantic evaluation was implemented using structured output with GPT-5
through the OpenAI API. The evaluator was run with temperature 1 and up to
eight parallel workers; failed requests were retried up to six times with
Supplementary Note 3.

\subsection{Characterization of knowledge-updating failures}

To characterize how incomplete knowledge updating manifested in generated
recommendations, responses were additionally classified according to a
predefined taxonomy of primary error modes. The taxonomy comprised fully
correct, minor omission, partial revision, stale prior retained, hybrid
old/new, wrong revision direction, critical qualifier omitted, unsupported
addition, wrong replacement, wrong population scope, wrong threshold, wrong
dose or schedule, wrong strength or certainty, and other substantive error.

Outdated-content retention was defined as the persistence of clinically
relevant guidance that had been superseded by the updated recommendation.
Hybrid old--new responses contained elements of both the superseded and
updated guidance in a manner that did not faithfully represent the current
recommendation. Partial revision indicated that the response implemented only
a subset of the substantive change, whereas omission categories captured
components of the updated recommendation that were absent from the generated
response. Critical qualifier omissions were distinguished from minor omissions
according to whether the omitted element materially changed the clinical scope,
conditions, strength, or interpretation of the recommendation.

Each response was assigned a single primary error category by the semantic
evaluator. The complete taxonomy and operational definitions are provided in
Table S5.

\subsection{Cross-model and evaluator robustness analyses}

To determine whether the recognition--incorporation gap was specific to the
primary model, we repeated the full-evidence evaluation across models spanning
multiple architectures, parameter scales, medical specialization, and
proprietary capability tiers. The evaluated models were Qwen3-4B, Qwen3-14B,
Qwen3-32B, Mistral-7B-Instruct-v0.3,
Mistral-Small-24B-Instruct-2501, Phi-4, Gemma-3-12B-IT,
MedGemma-4B-IT, MedGemma-27B-Text-IT,
Llama-3.1-8B-Instruct, Llama-3.1-70B-Instruct,
GPT-5.6 Luna, GPT-5.6 Terra, and GPT-5.6 Sol. All models were evaluated
on the same 1,709 recommendation-level revisions using the same
full-evidence task definition. For each model, we calculated revision
recognition, clinical correctness, full incorporation, and the
recognition--incorporation gap, defined as the percentage-point difference
between recognition and full incorporation.

We separately assessed robustness to the semantic evaluator. All 1,709
predictions from the primary Qwen3-4B full-evidence experiment were
independently re-evaluated using GPT-5.6 Sol as a second semantic evaluator.
The second evaluator applied the same endpoint definitions for revision
recognition, clinical correctness, and full incorporation. Agreement between
the primary and secondary evaluators was quantified using raw percentage
agreement and Cohen's $\kappa$ for each endpoint. We additionally compared
the recognition--incorporation gap estimated by the two evaluators.

\subsection{Revision-salience intervention}

We designed a revision-salience intervention to test whether incomplete
incorporation primarily reflected insufficient prioritization of updated
evidence over prior knowledge. The intervention used the same clinical
inputs and generation task as the standard full-evidence condition, but
strengthened both the system and user instructions to explicitly prioritize
updated guideline evidence over conflicting or outdated information in the
earlier recommendation. No additional clinical information was introduced
by the intervention.

We applied the intervention to the same 1,709 recommendation-level revisions
evaluated in the standard Qwen3-4B full-evidence condition, using otherwise
identical generation settings. Outputs were evaluated using the same semantic
evaluation protocol. We compared revision recognition, clinical correctness,
full incorporation, outdated-content retention, hybrid old--new responses,
and the recognition--incorporation gap between the standard and
revision-salience conditions. Because both conditions were evaluated on the
same revisions, differences were calculated at the paired case level.
The exact standard and revision-salience prompts are provided in
Supplementary Note 4.

\subsection{Clinician evaluation}

To independently assess the clinical validity of the automated semantic evaluation, we conducted a clinician assessment of 300 model responses from the ChronoEvo evaluation set. The assessment sample comprised 250 cases from the primary analysis set and 50 additional cases from an enrichment set, covering all 12 knowledge-revision types. The enrichment cases comprised replacement revisions ($n=20$), changes in recommendation strength or certainty ($n=10$), population expansions ($n=5$), population restrictions ($n=5$), recommendation removals ($n=5$), and threshold changes ($n=5$). The 300 cases represented 183 distinct guideline revision edges and 169 longitudinal guideline lineages. Given the enrichment of selected revision types, clinician-assessed proportions were reported descriptively for the validation sample and were not extrapolated to the full benchmark.

Two licensed physicians, specializing in orthopedic surgery and cardiac surgery, independently assessed each response. For every case, the clinicians were provided with the prior recommendation, updated guideline evidence, verified current recommendation, and model-generated response. Both clinicians were blinded to the automated semantic evaluation labels and to each other's initial assessments.

Each response was evaluated on three binary endpoints. Revision recognition indicated whether the response reflected the substantive direction of the guideline change. Clinical correctness indicated whether the response was clinically consistent with the updated guidance. Full incorporation required faithful integration of all clinically relevant components of the revision, including applicable population criteria, thresholds, doses, schedules, and recommendation strength. These endpoints were assessed separately, allowing a response to recognize a revision without fully incorporating it.

Following independent annotation, disagreements were resolved through consensus discussion to establish adjudicated clinician labels. Inter-clinician reliability was quantified using raw percentage agreement and Cohen's $\kappa$, calculated separately for each endpoint.

The adjudicated clinician labels were compared with the corresponding judgments of the primary automated semantic evaluator (GPT-5) on the same 300 responses. Agreement was quantified using raw percentage agreement and Cohen's $\kappa$ for each endpoint. We calculated the proportions of responses satisfying revision recognition, clinical correctness, and full incorporation, together with the recognition--incorporation gap, defined as the percentage-point difference between revision recognition and full incorporation. Finally, we cross-tabulated clinical correctness and full incorporation to determine the proportion of incompletely incorporated responses that were also judged clinically incorrect.

\subsection{Statistical analysis}

All analyses were conducted at the recommendation-revision level unless
otherwise specified. Continuous quantities are reported as means or mean
differences, and binary outcomes as proportions. Statistical tests were
two-sided with a nominal significance level of 0.05. Where multiple
comparisons were performed within an analysis family, $P$ values were adjusted
using the Benjamini--Hochberg false-discovery-rate (FDR) procedure.

For lexical revision gain, we summarized the mean per-instance gain and the
proportion of instances with positive gain ($G_i>0$). Uncertainty was
estimated using 10,000 bootstrap resamples, with percentile-based 95\%
confidence intervals. Comparisons of revision gain across semantic revision
types used bootstrap confidence intervals for mean differences and two-sided
permutation tests with 10,000 permutations, with pairwise $P$ values adjusted
using the Benjamini--Hochberg procedure. Randomized revision-gain analyses
used a fixed seed of 42.

For semantic outcomes, we first quantified heterogeneity across the 12
benchmark-assigned revision types using Pearson $\chi^2$ tests of
independence. Revision-type labels were used only for post-evaluation
stratification and were not provided to the semantic evaluator. As secondary
exploratory analyses, pairwise comparisons between revision types were
performed irrespective of the omnibus-test result, with $P$ values corrected
across all pairwise comparisons using the Benjamini--Hochberg procedure. The
association between revision type and the distribution of primary error
categories was similarly assessed using a Pearson $\chi^2$ test, with
Cramer's $V$ used to quantify effect size.

To assess whether revision-specific associations persisted after accounting
for measured characteristics of individual revisions, we fitted
multivariable binomial logistic-regression models for full incorporation of
the revision and retention of outdated content. Revision type was entered as
a categorical predictor, with addition as the reference category. Models
additionally adjusted for token-level lexical similarity between the earlier
and updated recommendations, the temporal interval between guideline
versions, the length of the earlier recommendation, the length of the updated
recommendation, and the length of the updated evidence. Continuous covariates
were standardized before model fitting. Results are reported as odds ratios
with 95\% confidence intervals. In the primary analysis, cluster-robust
standard errors were calculated at the guideline-revision-edge level to
account for dependence among multiple recommendation revisions arising from
the same guideline update. Benjamini--Hochberg correction was applied across
the 11 non-reference revision-type coefficients separately within each
outcome. As a sensitivity analysis for broader within-lineage dependence, the
models were repeated with cluster-robust standard errors calculated at the
reconstructed guideline-lineage level.

For cross-model analyses, each model was evaluated on the same set of 1,709
recommendation revisions. The recognition--incorporation gap was defined as
the difference between the revision-recognition rate and the
full-incorporation rate. Confidence intervals for revision recognition,
clinical correctness, full incorporation, and the recognition--incorporation
gap were estimated using 10,000 cluster-bootstrap resamples at the
guideline-revision-edge level. In each bootstrap replicate, revision edges
were sampled with replacement and all recommendation-level observations
belonging to a sampled edge were retained. Comparisons across model scales
were performed using paired cluster bootstrap after alignment by benchmark
instance; the same sampled guideline-revision-edge clusters were used for
both models within each bootstrap replicate. Differences were defined as the
larger or more capable model minus the smaller or less capable model and are
reported with percentile-based 95\% bootstrap confidence intervals.

The revision-salience intervention was evaluated in a paired design by
aligning standard and salience-enhanced responses for the same benchmark
instances. Revision recognition, clinical correctness, full incorporation,
outdated-content retention, and hybrid old--new responses were compared using
exact two-sided McNemar tests. Effect sizes were expressed as paired
differences in outcome proportions (salience-enhanced minus standard), with
95\% confidence intervals estimated using 10,000 paired cluster-bootstrap
resamples at the guideline-revision-edge level. In each bootstrap replicate,
revision edges were sampled with replacement and all paired
recommendation-level observations belonging to a sampled edge were retained.
The recognition--incorporation gap was calculated within each condition as
the recognition rate minus the full-incorporation rate and compared between
conditions using the same paired cluster-bootstrap procedure.

Robustness to semantic-evaluator choice was assessed by independently
re-evaluating the primary-model responses with GPT-5.6 Sol using the same
change-type-blinded evaluation framework. For each binary semantic endpoint,
we calculated raw inter-evaluator agreement and Cohen's $\kappa$. We
additionally compared the recognition--incorporation gaps obtained from the
two evaluators to assess whether the principal dissociation was sensitive to
evaluator choice.

\subsection{Semantic taxonomy of medical knowledge revisions}

Each verified recommendation-level revision was assigned a primary semantic
change type describing how the updated recommendation differed from the prior
recommendation. The taxonomy comprised 12 categories: addition, removal,
replacement, strength or certainty change, population expansion, population
restriction, threshold change, dose or schedule change, diagnostic change,
monitoring change, procedural change, and other.

Addition denoted the introduction of a clinically substantive recommendation
that was not present in the prior guideline, whereas removal denoted withdrawal
of previously recommended guidance without a corresponding replacement.
Replacement denoted substitution of prior guidance with substantively
different guidance addressing the same clinical focus. Strength or certainty
change captured revisions to the strength, confidence, or evidentiary
certainty of an otherwise related recommendation. Population expansion and
population restriction captured broadening and narrowing, respectively, of
the population to which a recommendation applied. Threshold change denoted a
change in a quantitative or categorical decision threshold. Dose or schedule
change captured changes in treatment dose, frequency, duration, or timing.
Diagnostic change captured revisions to diagnostic criteria, testing, or
diagnostic strategy; monitoring change captured revisions to surveillance or
follow-up recommendations; and procedural change captured changes in the
recommended use or conduct of clinical procedures. Revisions that were
clinically substantive but did not fit these prespecified categories were
assigned to the other category.

Each verified recommendation-level revision was assigned one primary semantic
change type. The initial category assigned during structured extraction was
independently reassessed during the verification stage and was corrected when
the verifier judged another category to better characterize the revision.
The resulting single category was retained as benchmark metadata for
post-evaluation stratified analyses and was not provided to either the generation model or the semantic evaluator. Operational definitions for all 12 categories are provided in 
Supplementary Note 5.

\section{Data Availability}

The source TempoMed-Traj dataset is publicly available from its original repository (https://huggingface.co/datasets/anonymous2509/TempoMed-Traj). Source clinical-guideline documents are not redistributed where restricted by the terms or licensing conditions of the original publishers or repositories. The code and provenance-based reconstruction procedures for ChronoEvo will be publicly available through Zenodo upon publication.

\section{Code Availability}


Code for constructing ChronoEvo and reproducing the experiments and analyses reported in this study is available to editors and reviewers through a Zenodo preview link.

\section*{Acknowledgements}
This work was supported by the National Center for Complementary and Integrative Health, United States [grant numbers R01AT009457, U01AT012871]; the National Institute on Aging, United States [grant number R01AG078154]; the National Cancer Institute, United States [grant number R01CA287413]; the National Institute of Diabetes and Digestive and Kidney Diseases, United States [grant number R01DK115629]; the National Institute on Minority Health and Health Disparities, United States [grant number R21MD019134]; and the Food and Drug Administration, United States [grant number U01FD008720]. The funders had no role in the study design, data collection and analysis, decision to publish, or preparation of the manuscript.

\section*{Author contributions}
M.Z. and R.Z. conceived and designed the study.
M.Z. developed ChronoEvo, implemented the computational framework,
conducted the experiments, analyzed the results, and drafted the manuscript.
M.Z., Z.Z., K.Y. and R.Z. contributed to scientific discussions
on the study design, experimental findings, and interpretation of the results.
R.Z. supervised the research, provided scientific guidance, and acquired funding.
All authors contributed to reviewing and revising the manuscript
and approved the final submitted version.

\section*{Competing Interests}
The authors declare no competing interests.


\bibliography{sn-bibliography}

\clearpage
\setcounter{figure}{0}
\setcounter{table}{0}
\renewcommand{\thefigure}{S\arabic{figure}}
\renewcommand{\thetable}{S\arabic{table}}

\end{document}